\documentclass{ieeeaccess}
\usepackage{cite}
\usepackage{amsmath,amssymb,amsfonts}
\usepackage{algorithmic}
\usepackage{graphicx}
\usepackage{textcomp}
\def\BibTeX{{\rm B\kern-.05em{\sc i\kern-.025em b}\kern-.08em
    T\kern-.1667em\lower.7ex\hbox{E}\kern-.125emX}}
\begin{document}
\history{Received July 19, 2021, accepted September 6, 2021, date of publication September 23, 2021, date of current version September 30, 2021.}
\doi{10.1109/ACCESS.2021.3115082}

\title{Object Permanence Through Audio-Visual Representations}
\author{\uppercase{Fanjun Bu}\authorrefmark{1},
\uppercase{Chien-Ming Huang \authorrefmark{2}}.
\IEEEmembership{Member, IEEE}}
\address[1]{Department of Computer Science, Cornell University, Ithaca, NY 14850 USA (e-mail: fb266@cornell.edu)}
\address[2]{Department of Computer Science, Johns Hopkins University, Baltimore, 
MD 21218 USA (e-mail: cmhuang@cs.jhu.edu)}


\corresp{Corresponding author: Fanjun Bu (e-mail: fb266@cornell.edu).}

\begin{abstract}
As robots perform manipulation tasks and interact with objects, it is probable that they accidentally drop objects (e.g., due to an inadequate grasp of an unfamiliar object) that subsequently bounce out of their visual fields. To enable robots to recover from such errors, we draw upon the concept of \emph{object permanence}---objects remain in existence even when they are not being sensed (e.g., seen) directly. In particular, we developed a multimodal neural network model---using a partial, observed bounce trajectory and the audio resulting from drop impact as its inputs---to predict the full bounce trajectory and the end location of a dropped object. We empirically show that: 1) our multimodal method predicted end locations close in proximity (i.e., within the visual field of the robot’s wrist camera) to the actual locations and 2) the robot was able to retrieve dropped objects by applying minimal vision-based pick-up adjustments. Additionally, we show that our method outperformed five comparison baselines in retrieving dropped objects. 
Our results contribute to enabling object permanence for robots and error recovery from object drops.
\end{abstract}


\titlepgskip=-15pt

\maketitle

\section{Introduction}
\label{sec:introduction}
\begin{figure}
\centering
  \includegraphics[width=3.5in]{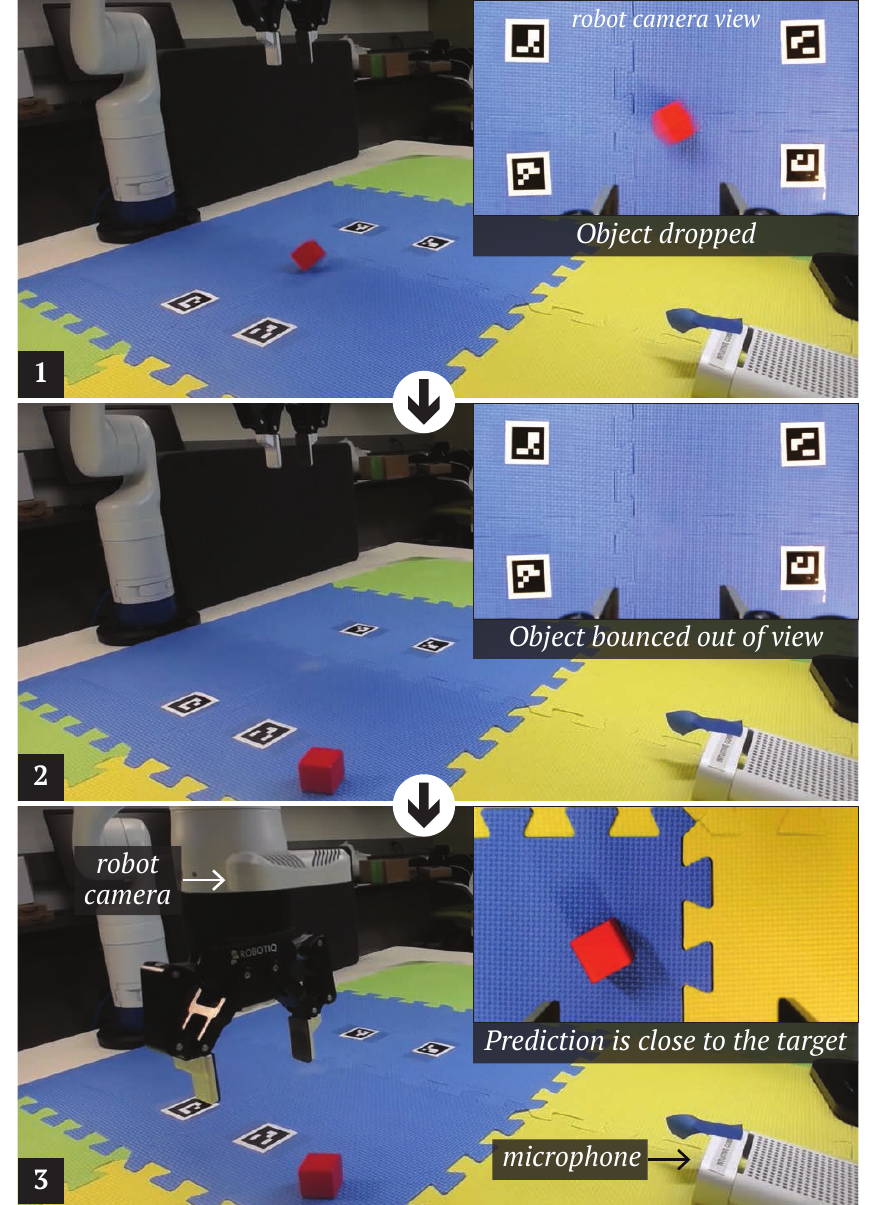}
  \caption{We explore object permanence through audio-visual representations and contextualize our exploration in retrieving dropped objects that bounce out of its visual field.}
\label{fig:setup}
\end{figure}

\PARstart{W}{e} all drop objects, from car keys to pens or utensils. When an object is dropped and bounces out of sight, people retrieve the object by estimating its landing position. This ability to retrieve objects that are outside of visual fields is based on an understanding of \emph{object permanence}  \cite{piaget1952origins}---objects remain in existence even when they may not be visible. As robots increasingly interact with human-made objects, they are bound to drop objects. Drops can be frequent not only due to the challenging nature of robot grasping \cite{saxena2008robotic}, imperfect sensing, complex human-made objects in unstructured human environments, and computationally intensive pose estimations, but also because robots may accidentally drop objects during manipulation. This paper addresses the problem of \emph{how robots may recover from dropping objects}, focusing on how to retrieve objects that bounce out of the robot's visual field by estimating bounce trajectories and object locations (Fig. \ref{fig:setup}). 

Drawing inspiration from humans' ability to locate objects using multiple sensing modalities, we developed a multimodal deep network that takes as input a partial, observed trajectory and impact sound to predict an object’s bounce trajectory and end location. Our multimodal network encodes modality-specific features from visual and auditory channels and uses observed partial trajectories to adjust audio-driven spatial predictions enabled by a microphone array. To illustrate our approach to object permanence, we used a simple wooden object and compared our approach to physics-driven baselines and baselines informed by previous state-of-the-art models. We additionally explored the possibility of generalizing our approach to a different object and dropping height. 
However, rather than focusing on developing a universal model capable of predicting end locations for objects made of various materials under a diversity of conditions, our goal in this paper is to demonstrate the utility of leveraging audio input and incomplete visual data to achieve object permanence. 

Our contributions include: 1) a human-inspired multimodal approach to object permanence, 2) a demonstration of how a robotic manipulator may use the multimodal approach to recover from object dropping, and 3) an open dataset of object drops for the robotics community.
Next, we briefly highlight relevant prior research that motivates this work.

\section{Background and Related Work}
\subsection{Object Permanence in Robotics}
The concept of object permanence originates from the fields of psychology and child development, and has motivated research in developmental robotics \cite{luo2007rethinking}. 
Prior research on object permanence in robotics has focused on realizing the concept of object permanence to enable robots to understand the effects of their own actions and facilitate their interactions with people. For example, object permanence has been explored to enable perspective taking and to support situated dialog between humans and robots \cite{roy2004mental, hsiao2003coupling}.  It has also been studied as a mechanism for self-monitoring, allowing robots to distinguish self-generated movements from external movements and thus infer the effects of their own actions. Such monitoring mechanisms can help robots form mental representations of objects that were occluded by their own actions \cite{bechtle2015first, bechtle2016sense, lang2018deep}. In contrast to previous research, the present work focuses on investigating how object permanence may be realized by modeling audio-visual inputs and used by a manipulator to retrieve objects that bounce out of its view.

\subsection{Trajectory Prediction}
The problem of trajectory prediction has been widely investigated to enable robot autonomy. For instance, a growing body of research focuses on modeling and predicting pedestrian trajectories to enable mobile robots to navigate in a safe and socially appropriate manner among people (e.g., \cite{gupta2018social, katyal2020intent, alahi2016social, tanaka2012motion, rudenko2020human}). Recent works have also investigated vehicle trajectory prediction in an effort to develop autonomous vehicles (e.g., \cite{deo2018convolutional, houenou2013vehicle}). The common property shared by pedestrian trajectories and autonomous vehicle trajectories is that both are goal-oriented. When the trajectories are not driven by intentional agents but purely laws of physics, physics-based and dynamics models are commonly used (e.g., \cite{chen2010dynamic, huang2011trajectory}). For example, 
neural network models have been proposed to infer physical parameters from images, capturing the physical interactions between a foam ball and different type of surfaces, and predict post-bounce trajectories of the foam ball for a certain time interval \cite{purushwalkam2019bounce, InnamoratiEtAl:DynamicBounce:ICCV:2019}. In our work, we focus on the prediction of bounce trajectories and end locations of dropped objects with the goal to enhance robot autonomy in events of unintended object drops. 

\subsection{Audio-Visual Joint Reasoning}

Visual and auditory signals are typically correlated in common daily activities. This insight has inspired learning sound representations from visual signals  \cite{aytar2016soundnet} and led to various explorations of object/scene recognition through audio-visual joint reasoning. 
For example, cross-modal models, leveraging audio-visual correspondence, have been used to map audio and visual data to a shared space and locate the sound source in visual data \cite{arandjelovic2017look, arandjelovic2018objects}. 
With audio-guided visual attention, a multimodal residual network has shown success in audio-visual event localization in video segments \cite{tian2018audio}. Audio-visual correspondence also allows audio signals to serve as supervision for visual learning \cite{owens2016ambient}. 

In the context of robotics and situated interactions, joint reasoning using audio-visual inputs supports target localization, tracking, and hence navigation (e.g., \cite{Wilson2020, chau2019audio, ban2019variational, pleshkova2016audio}). For example, a deep neural network that utilizes both vision and auditory inputs outperforms conventional methods that rely on single modalities in object tracking \cite{Wilson2020}. 
Audio-visual fusion has also been used to enhance speaker tracking \cite{shivappa2010audio, 8656587}. A simultaneous localization and mapping (SLAM) framework that uses both audio and visual inputs has been shown to localize both human speakers and the observer effectively \cite{chau2019audio}. A variational expectation-maximization algorithm that takes audio and visual signals as inputs is able to track the location of multiple speakers in the scene \cite{ban2019variational}. Rather than focusing on tracking, this work focuses on predicting a partially unobservable trajectory that resulted from an object drop using the observed portion of the trajectory and a complete impact sound recording.

\begin{figure*}[t!]
\centering
\includegraphics[width=\linewidth]{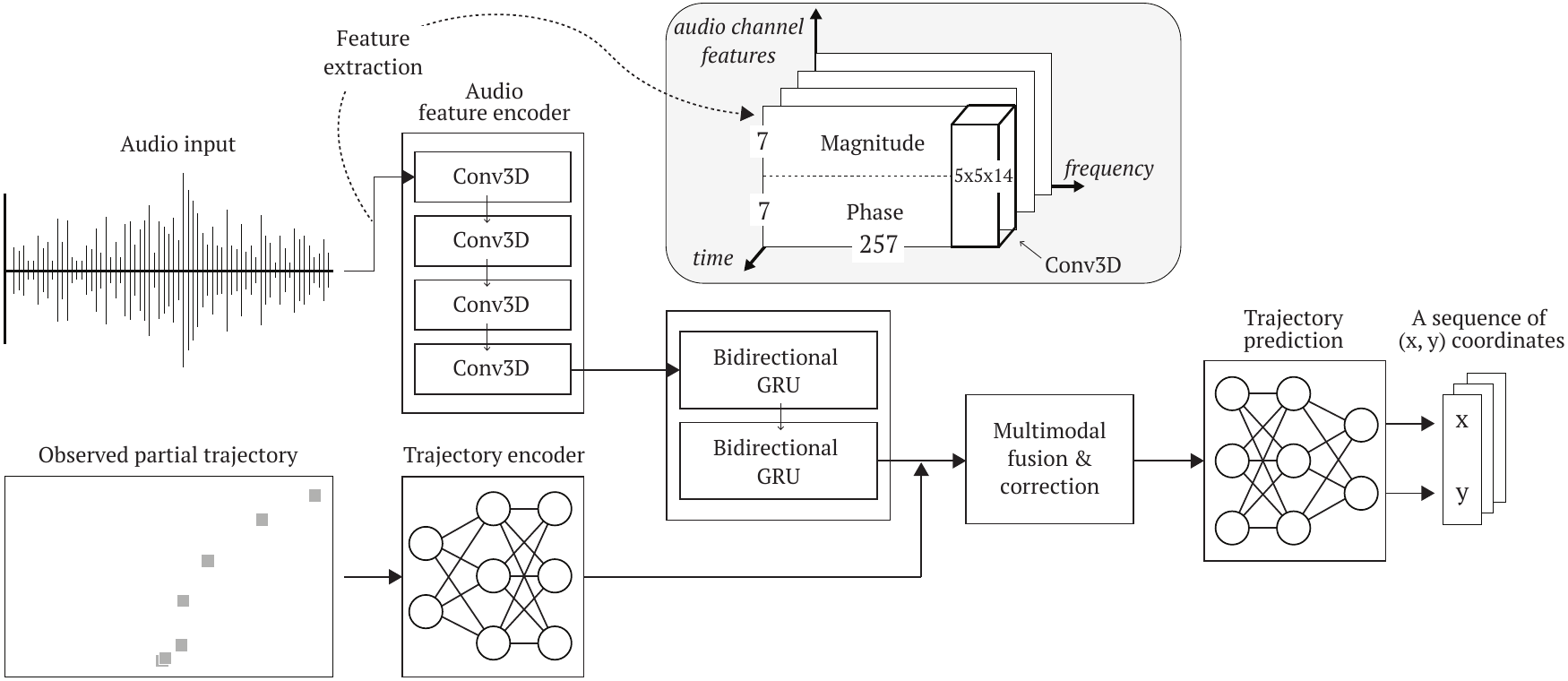}
\caption{Structure of our multimodal neural network.}
\label{fig:model}
\end{figure*}

\section{Problem Statement}
Our goal is to enable robots to recover from inadvertent object drops.
In this work, we consider a robotic manipulator with a wrist camera and that does not rely on external visual sensing capability for error recovery, and moreover, we focus on erroneous situations in which objects bounce out of the robot's visual field.
To contextualize our investigation, we experimentally let the robot drop a wooden block and collected its partial trajectory, as observed through its wrist camera, and the corresponding impact sound due to the dropping. Our primary technical problem is the recovery of the full bounce trajectory and end location of the dropped object given the partial trajectory and impact sound.

\section{Multimodal Network for Object Permanence}
We developed a new network architecture\footnote{{Our code is available at: \underline{https://intuitivecomputing.jhu.edu/openscience.html}}} (Fig. \ref{fig:model}) that takes as inputs an observed partial trajectory and a complete audio recording, and outputs the object's bounce trajectory and its end location with respect to the robot. 

\subsection{Feature Extraction}
\subsubsection{Audio} Similar to the approaches proposed in SELDnet \cite{adavanne2019localization} and DOAnet \cite{adavanne2018direction}, we extracted magnitude and phase components from spectrograms of each channel of the microphone array. 
The magnitude and phase components were then stacked along the channel dimension and  treated as a $M/2$ by \(2\times C\) image, where $M$ is the window length of the Fourier transformation and $C$ is the number of channels (Fig. \ref{fig:model}). After processing the whole audio sequence with 50\% overlap on window size, the shape of the audio data was \(T \times M/2 \times 2C\), where $T$ is the number of data points in the time dimension. The parameters used for our feature extraction followed those in SELDnet \cite{adavanne2019localization} and DOAnet \cite{adavanne2018direction}.

\subsubsection{Vision} Observed trajectories were represented as NumPy arrays with shape $(65, 2)$ after being extracted from series of images using a color tracking algorithm (detailed in Section \ref{sec:exp1}).

\subsection{Audio Feature Encoder Module}
As informed by SELDnet \cite{adavanne2019localization}, we encoded audio features through convolutional recurrent neural networks.
To learn inter-channel and intra-channel features across time, we used four 3D convolutional layers with kernel size $(k, k, c)$ to compute our audio representations. $(k, k)$ is the kernel size in the time-frequency dimension, which aims to capture correlations within the channel across time; $c$ is the total number of channels, which allows for finding correlations across all available channels. 3D max pooling layers were added between convolutional layers to reduce the dimension along the frequency axis. 

The encoded audio representations were then passed to two bidirectional Gated Recurrent Units (GRUs) (input dimension = 64, hidden dimension = 64, number of layers = 1), seeking to further encode possible connections in the time domain. 

\subsection{Trajectory Encoder Module}
The trajectory encoder module is a three-layer multilayer perceptron (MLP) that maps the \(65\times2\) partial observed trajectories from the robot's wrist camera to a higher-dimensional feature vector that matches the shape of the output of the bidirectional GRUs (\(T \times 128\)).

\subsection{Mutimodal Fusion \& Correction Module}
The audio representation (\(T \times 128\)) and the visual feature vector (\(T \times 128\)) are stacked into one 2D vector \((T \times 256)\), which is subsequently fed into a two-layer MLP. The audio representation reflects the complete trajectory, while the visual representation describes the observed, partial trajectory. Consequently, the visual representation serves as a mechanism to provide corrective adjustments to the location information embedded in the audio representation. The output of the fusion and correction module \((135 \times 256)\) encodes each $(x, y)$ coordinate in a 256-dimensional vector, with 135 being the time length for complete trajectories.

\subsection{Trajectory Prediction Module}
The trajectory prediction module is a three-layer MLP (hidden layer size: $(64, 16)$) that decodes the 256-dimensional vector representation from each time step and outputs a corresponding $(x, y)$ location. We note that our prediction includes the full trajectory so as to ensure robustness against missing frames. 

\section{Experiment}
\label{sec:exp1}
In this section, we describe an experiment that sought to evaluate the effectiveness of our multimodal model in predicting the object bounce trajectory and end location based on a partially observed trajectory and impact sound. 

\subsection{Task and Data Collection Setup} 
Our experimental task involved a robot picking up a red wooden cube (3 cm x 3 cm x 3 cm) and releasing it from the same height (0.3 meters above the table surface) repeatedly. There was no systematic manipulation of object release across trials. Observed differences in trajectories and impact sounds were results of gripper friction and natural physics.
Our data collection setup is shown in Fig \ref{fig:setup}. We used a Kinova Gen3 robot, which has a built-in RGB-D wrist camera (\textit{robot camera}). A Microsoft Azure DK camera was used to collect impact sounds; we note that we only used its embedded microphone array (seven channels; 48k sampling rate). The microphone was positioned to reduce noise from the robot motors. A RealSense D435 camera was mounted on the ceiling (\textit{ceiling camera}) to collect the ground truth videos, which included object trajectories outside of the robot camera's visual field. Both the robot and ceiling cameras had a data rate of 30 frames per second.

\begin{figure*}[t]
\centering
\includegraphics[width=\linewidth]{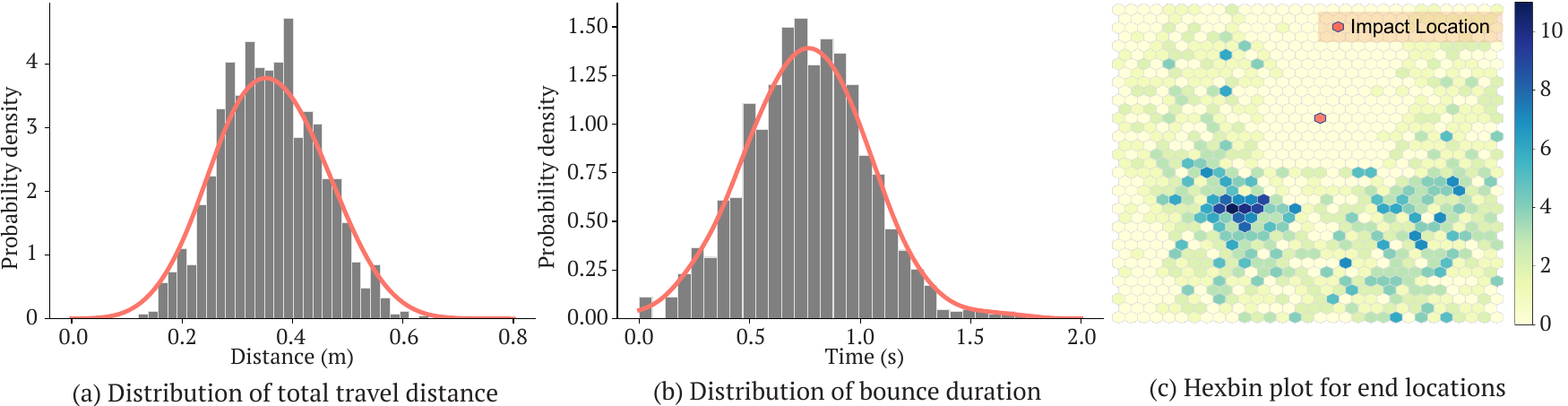}
\caption{(a) This histogram presents the distribution of travel distance after the initial impact between the cube and surface. The red curve shows the estimated Gaussian distribution. (b) This histogram depicts the distribution of travel duration after the initial impact between the cube and surface. (c) This hexbin plot shows the density of end locations in the 2D field as viewed from the ceiling camera. Each cell represents the number of times the cube ended up in that cell. The robot was placed at the top facing down. We only visualize the end locations that were outside the initial visual field of the robot's wrist camera. Thus, the rectangular empty space in the middle corresponds roughly to the size of the visual field.}
\label{distr}
\end{figure*}

\subsection{Experimental Data}
\subsubsection{Data Collection and Processing}
In each trial, the robot camera and the ceiling camera published RGB images separately to different ROS topics. 
To synchronize all of the recording devices (two cameras and a microphone array), two processes were run right before the robot released the object; the first process listened to both ROS topics and saved images in a rosbag file, whereas the second process recorded audio data using the PyAudio package. All processes ran in parallel for three seconds.

To extract object trajectories, we wrote a simple color tracking program using OpenCV. 
In particular, we defined the range for the color red in HSV space and masked out the contours of red objects. The center of the object contour was recorded to represent the current location of the object.

Each camera, in practice, captured only a portion of the complete trajectory. The robot camera captured the trajectory up to the moment when the cube bounced outside of its view (\textit{observed trajectory}). The ceiling camera was initially blocked by the robot gripper and therefore was not able to observe the release of the cube. However, the ceiling camera was able to keep track of the cube after the cube was outside of the robot camera's visual field (\textit{post trajectory}). 

To map coordinate frames between the two cameras, four ChArUco markers were used such that they were initially visible to both the ceiling camera and the robot camera, allowing the post trajectory to be transformed to the coordinate frame of the robot camera. After this transformation, the overlapping portion between the observed trajectory and the post trajectory was removed from the post trajectory.

We kept the last 65 coordinates recorded in the observed trajectory and the first 70 coordinates recorded in the post trajectory to keep the model input size fixed. The numbers 65 and 70 were chosen to ensure the complete capture of each trajectory. 
If the length of an observed trajectory was greater than 65, the beginning of the observed trajectory could be removed without losing information since the cube was still in the air while the first few coordinates were recorded. If the length of a post trajectory was longer than 70, the last few coordinates could be removed without losing information since the cube always stopped moving during the three-second movement window, leaving the last few data points the same. We removed cases in which the cube did not stop moving within 3 seconds; in most of these cases, the cube fell off the table in our experimental setup. For similar reasons, if the length of an observed trajectory or a post trajectory were less than 65 or 70, we could prepend or postpend the first or last data point, respectively, to keep the input size of our network fixed. After the above processing steps, a complete trajectory was formed by concatenating an observed trajectory with its corresponding post trajectory.

We note that our audio and trajectory data were not necessarily aligned. The number of trajectory data points that was removed depended on factors such as missing frames and travel direction, and was independent of the corresponding audio data. We did not trim or pad audio data accordingly to keep two data sequences aligned.

\subsubsection{Dataset}
A total of 1404 trajectories\footnote{Data is available at: \underline{https://doi.org/10.7281/T1/EP0W7Y}} was included in this experiment \cite{dataset}. All included data points contained full trajectories from the ceiling camera's point of view; instances where the experimental cube rolled out of the ceiling camera's view (most likely fell off the table) or never left the initial visual field of the wrist camera were excluded. We also excluded the data points where the cube bounces backward and collide with the robot. 

Each data point consists of a three second long audio recording and two NumPy 2D arrays that contain the pixel location of the cube in the image frames from the robot camera and the ceiling camera. An observed trajectory from the robot's wrist camera is a NumPy array with shape (65, 2); a complete trajectory from the ceiling camera is a NumPy array with shape $(135, 2)$. Each row in the trajectories is the $(x, y)$ coordinate of the object in the image plane. All trajectories were normalized for our proposed model, and the initial locations of the trajectories were set as the origin using the equation: \( trajectory = trajectory[0] - trajectory\).

Fig. \ref{distr} visualizes characteristics of our dataset in terms of bounce duration, distance traveled, and resulting end locations. We calculated the duration of object bounces by computing the difference between the first peak and last peak in audio wave data. We also computed the geodesic distance traveled by the object (i.e., Euclidean distance between initial impact location and the cube's end location). 
The distributions of bounce duration and geodesic distance can be modeled as Gaussian distributions through kernel density estimation. The parameters for the distributions were dependent on factors including object properties (e.g., shape, size, material) and surface properties. 
Fig. \ref{distr} (c) shows the distribution of the object's end location in the ceiling camera's visual field.

\subsection{Evaluation Baselines}
We evaluated our proposed method against five baselines: two simple uni-modal baselines, two additional uni-modal baselines inspired by state-of-the-art (SOTA) methods, and one multimodal baseline that combined the two SOTA models. We provide detailed descriptions below.

\subsubsection{B1: Linear Extrapolation of Visible Trajectory (simple vision-only baseline)}
One simple baseline for estimating the end location of a dropped object is to project its trajectory linearly based on the last seen location ($P_f$) before the object bouncing out of view, the corresponding velocity at that location, and the remaining time to travel. 
We can compute the velocity ($v$) at which the cube leaves the visual field by multiplying the coordinate difference between the last two observed frames and the inverse of sampling frequency (1/30). 
From impact sound, we can compute the duration of bouncing ($t$) after the cube leaves the visual field. Assuming that the acceleration is constant and that velocity linearly decreases to zero, we can then estimate the end location of the cube using the following equation:
\begin{equation} P_f + vt + \frac{1}{2}at^2, \text{where } a = \frac{0-v}{t}\end{equation}

\subsubsection{B2: Estimation of Audio Signal Delay (simple audio-only baseline)} Given the microphone arrays used in our data collection, localization estimation through sound can be solved by leveraging signal delay across microphone channels \cite{fan2010localization}. 
We used convolutional layers to extract delay information across channels as our audio and visual data were not perfectly synchronized due to noises and difference in sampling frequency. 
Specifically, the output of our 7-D microphone array was treated as a \(T\) by 7 image, where \(T\) denotes time. The image was fed through three convolutional 2D layers with kernel sizes (16, 7), (8, 7), and (4 ,7). Each convolutional layer was followed by a 2D batch normalization layer and a 2D pooling layer. The output of the final convolutional layer was reshaped by a multi-layer perceptron. 

\subsubsection{B3: Social GAN-lite (SOTA vision-only baseline)}
Most trajectory prediction models utilize state-of-the-art recurrent neural network architectures.
This baseline was motivated by one such architecture: the Social GAN \cite{gupta2018social}, which utilizes long short-term memory (LSTM) \cite{hochreiter1997long} in predicting human movement trajectories. 
In particular, we embedded each 2D coordinate of the observed trajectory in a 64-dimensional vector. Then, an encoder (implemented using LSTM, input dimension = 64, hidden dimension = 64, number of layers = 1) took the 65 embedding vectors as inputs and outputted its final hidden state. A decoder (implemented using LSTM, input dimension = 64, hidden dimension = 128, number of layers = 1) took the 64-dimensional embedding of the last 2D location coordinate of the observed trajectory and the final hidden state of the encoder as inputs and recursively predicted the difference between consecutive locations sequentially for the 70 unseen trajectory positions, which we normalized. Our choice of parameters was consistent with that in Social GAN \cite{gupta2018social}. Note that the decoder predicted recursively starting at the end location of the observed trajectory. Thus, we needed to include the observed trajectory in our predicted complete trajectory. This is different from our proposed multimodal method, where we did not take the observed trajectory as exact in the prediction.

\subsubsection{B4: SELDnet-lite (SOTA audio-only baseline)}
The modal used for this audio baseline was inspired by SELDnet \cite{adavanne2019localization} and mirrored our multimodal network, except that it did not include the vision input, vision encoder, and fusion components. The audio representation produced by the bidirectional GRUs was passed to the trajectory prediction network directly. Conceptually, this baseline addressed the sound localization problem without vision-based corrective adjustments.

\subsubsection{B5: Combination of Social GAN-lite and SELDnet-lite}
In addition to four uni-modal baselines described above, we created a multimodal baseline that combined the Social GAN-lite and SELDnet-lite models.
Specifically, the output from the Social GAN-lite's encoder and the output from the SELDnet-lite's bidirectional GRU were fused through a fully connected linear layer and then reshaped to match target shape.

\subsection{Model Training} 
We split our 1404 data samples into training, validation, and test sets using an 8:1:1 ratio. We trained all models for 30 epochs with the Adam optimizer using a learning rate of $1e-4$ and a momentum parameter of 0.9. No early stopping was applied, and model selection was performed using the validation set.

\subsection{Evaluation Metrics} 
To compare model performance, we employed two evaluation metrics, focusing on whether or not the robot was able to retrieve the dropped objects successfully and on how accurate the predictions of the end locations were. Below, we describe these metrics in detail. 

\subsubsection{Task Success Rate}
In practice, the robot is able to reposition itself using a simple vision-based alignment method to retrieve the object when it is in the camera view (Fig. \ref{fig:setup}). Therefore, we considered a prediction as successful if the dropped object was visible to the robot's wrist camera when the robot was located at the end of the predicted trajectory.  We defined task success rate as the number of trials with a successful prediction divided by the total number of trials.
\begin{equation} \text{Task Success Rate} = \frac{ \text{Number of Successful Trails}}{\text{Total Number of Trials}}\end{equation}

\subsubsection{Target Displacement Error}
Target displacement error represents the distance (in centimeters) between the end location of a predicted bounce trajectory and the ground truth end location.

\subsection{Results}
\begin{figure*}[t]
\centering
  \includegraphics[width=\linewidth]{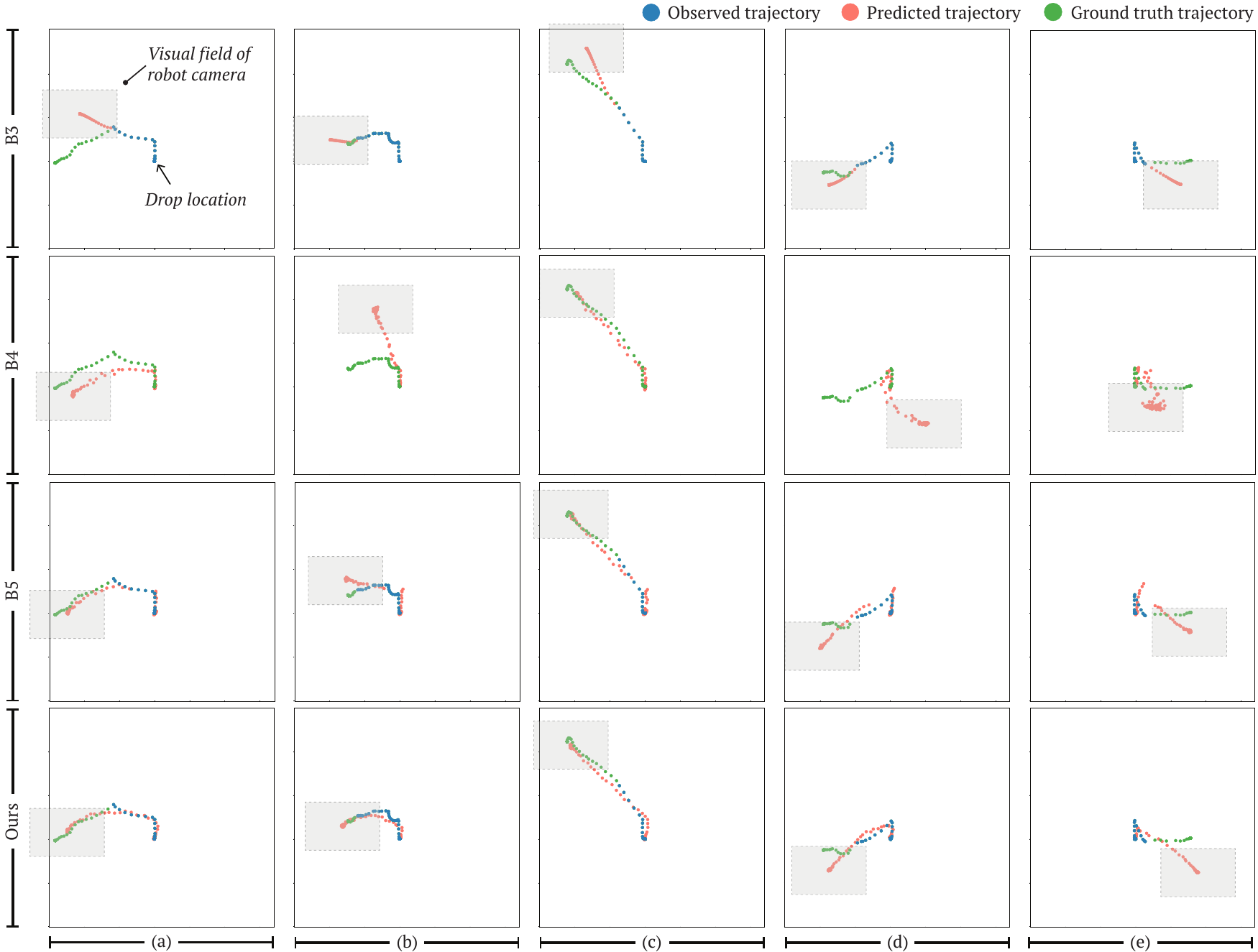}
\caption{Sample predictions from baselines (B3, B4, B5) and the multimodal model. Each column shows prediction results from the same trial.}
\label{ExampleTraj}
\end{figure*}

We conducted independent two-tailed t-tests to compare the target displacement errors from our multimodal model and the baseline models. Table \ref{table1} summarizes the results of the model performance on the test dataset. Figure \ref{ExampleTraj} shows samples of trajectory predictions. 
Overall, our multimodal network outperformed all baselines in all metrics, suggesting that the learned audio-visual representation captured meaningful location information from both modalities.

Linear extrapolation of visible trajectory (B1) made reasonable predictions when the unobservable portion of the trajectory was mostly linear. However, the majority of the collected trajectories were non-linear, influenced by the object and surface properties (Fig. \ref{ExampleTraj}).
The Social GAN-lite model (B3) also tended to predict linear trajectories, although it appeared that the Social GAN-lite model may have relied on two sources of information that were implicitly contained in the partial observations for prediction: 1) the distance between consecutive observed locations (capturing bounce speed) and 2) directional change in consecutive observed locations. When a bounce trajectory was close to linear, this vision model worked reasonably well (Fig. \ref{ExampleTraj} (c)). 

Through estimation of audio signal delay, the B2 model seemed to capture non-linearity and the general shape of a bounce trajectory. However, the visualization of output trajectories (Fig. \ref{ExampleTraj}) indicated that the model failed to capture the sequential property of input data, as predicted trajectories were noisy and scattered. 
In contrast, the SELDnet-lite model (B4) seemed to be able to capture non-linearity and enforce the sequential property of the trajectory at the same time. 
We believe that the phase and magnitude information extracted from the seven-channel microphone array carried more nuanced information that the 2D visual trajectories did not represent, and that the recurrent neural network component in B4 effectively maintained the sequential property of trajectories. However, early errors (offsets) propagated through the model and influenced later predictions. As shown in Fig. \ref{ExampleTraj} (d) and (e), the offsets accumulated and caused consequential displacement error in the end.

B5, a multimodal baseline that combined B3 and B4, showed enhanced performance in both task success rate and target displacement error. However, simply combining the state-of-the-art uni-modal models was not sufficient. In B5, audio and visual information were both processed sequentially and appeared to interfere with, rather than complement, each other, resulting in discontinuous trajectories (Fig. \ref{ExampleTraj} (e)). Additionally, as we will discuss in the next section, B5 did not generalize well to new bounce dynamics (Table \ref{table2}).

Our proposed multimodal model was able to utilize information from both audio and vision modalities effectively. Specifically, instead of a recurrent neural network (LSTM), a multi-layer perceptron was applied to partially observed trajectory data and able to correctively adjust early prediction offsets and reduced the error propagation.

In Fig. \ref{ExampleTraj} (b), (d) and (e), the accumulation of early offsets led B4's predictions away from the ground truth. In contrast, our multimodal model was able to utilize the observed information to reduce the errors from the audio modality and produced more accurate predictions. However, the model might not be able to recover from significant early prediction errors from the audio input (potentially due to surrounding noise), as shown in Fig. \ref{ExampleTraj} (e). 

\begin{table}[t]
\centering
    \caption{Results of the baselines and multimodal models on the test dataset.}
    \includegraphics[width=3.4in]{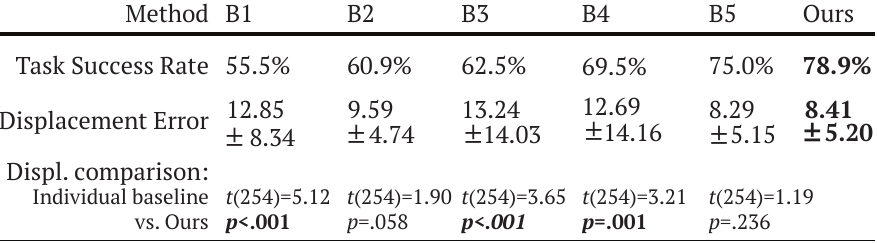}
    \label{table1}
\end{table}

\begin{table}[t]
    \centering
    \caption{Results of the finetuned baselines and multimodal model using Dataset H and Dataset T on their respective test datasets.}
    \includegraphics[width=3.4in]{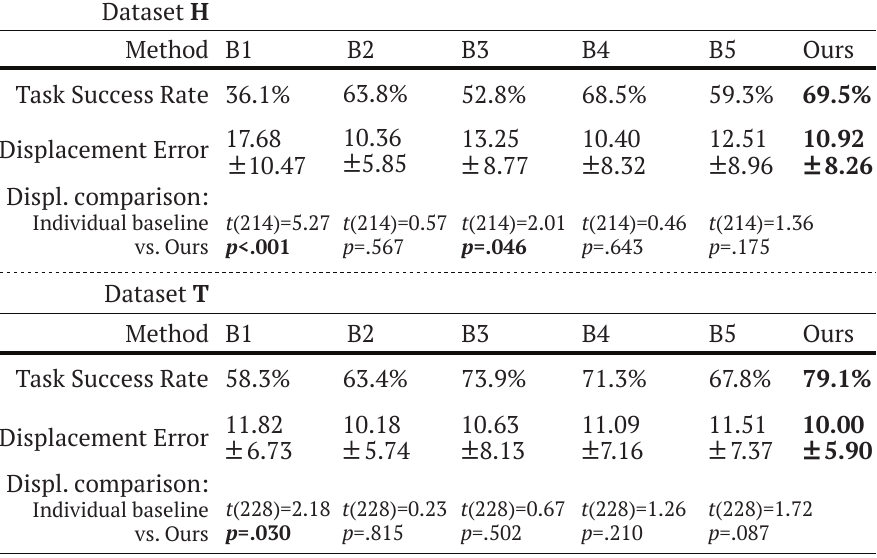}
    \label{table2}
\end{table}

\subsection{Additional Exploration}
While our goal in this paper is not to develop an omnipotent network that predicts well in every possible scenario, we wanted to explore the generalizability of our audio-visual representation. To this end, we collected two additional small datasets: releasing the original object (cube) from a different height (Dataset H, size 108) and releasing a different object from the original height of 0.3 meters (Dataset T, size 115). Specifically, for Dataset H, the cube was released 5 cm higher than the height used in the original experiment. For Dataset T, we used a triangular wooden block to replace the cube. The triangular block can be thought of as the original cube cut in half along the diagonal. These datasets include novel bounce dynamics that were not represented in the original dataset. 
In Dataset H, the cube made the initial impact with more energy, which resulted in longer and more non-linear trajectories. 
In Dataset T, the triangular block was lighter in weight and non-symmetric. Therefore, the magnitude of impact audio was smaller, and conversely noise was more evident. Trajectories in Dataset T also tended to be non-linear.

To explore generalizability, we finetuned all baselines and our proposed model using Dataset H and Dataset T separately, with a much smaller learning rate ($1e-5$) and 0 momentum value for 3 epochs. We performed a 5-fold cross-validation on each dataset to account for small data size.
Table \ref{table2} describes the results of our exploratory examination. These results suggest that a small set of data was sufficient in finetuning our multimodal modal to achieve reasonable performance. Note that both the audio baseline and multimodal model performed reasonably well, suggesting the importance of audio data in providing indicative information.

\section{Robot Demonstration}
Fig. \ref{fig:setup} illustrates a manipulator using our multimodal method to locate and retrieve a dropped object. A full demonstration can be seen in our supplementary video\footnote{\underline{https://youtu.be/Rj-ZZf3r4g8}}. 
To allow for retrieval actions, we transformed model predictions in the robot camera frame to the world frame. Though predictions were not perfect, as long as the object was present in the robot's camera view, the robot was able to use a simple vision-based method to reposition itself for object grasping.

\section{Discussion}
Object permanence is crucial for autonomous robots to interact with objects and operate in human environments robustly. In this project, we explore object permanence through object localization and retrieval in the context of dropped objects bouncing out of a robot's visual field. We developed a multimodal neural network that combines a partial, observed bounce trajectory and the audio resulting from the drop impact to predict the full bounce trajectory and the end location of a dropped object. Our results show that our approach outperformed various baseline methods. 

\subsection{Object Permanence for Enhancing Robot Operations}
The ability to estimate where dropped objects may be is important in enhancing robot operations. As an example, our multimodal model can be used to let robots recover from accidentally dropping objects efficiently. Rather than relying on external sensing or heuristics-based search, our lightweight model is able to provide reasonable estimations for object retrieval.
This ability to estimate object locations also provides robots with a sense of action feasibility. If an object is too far to reach or in a position that the robot cannot find a feasible motion plan to reach, the robot may instead ask for human assistance, minimizing unnecessary grasping attempts. Overall, this ability may be used to foster the fundamental skill of knowing when to ask for help.

\subsection{Limitations and Future Work}
The task setup used in this work was experimentally controlled. Future work should explore task settings in natural human environments, diverse dropping scenarios (e.g., accidental drops during robot motions), and a variety of experimental objects (e.g., everyday objects). Future work also needs to explore different audio-visual representations that can encode richer nuances in object properties and physical impact. For instance, both a higher release point and a heavier object can result in an increase in magnitude. In addition to different representations, future work should also consider other sensing modalities that might contribute to the formation of object permanence understanding. Lastly, more research is need to study object permanence in the context of human-robot collaboration in which objects are jointly used, manipulated, and shared.

\section*{Acknowledgments}
We would like to thank the Johns Hopkins University Institute for Assured Autonomy for supporting this work.

\bibliographystyle{IEEEtran}
\bibliography{access}

\begin{IEEEbiography}[{\includegraphics[width=1in,height=1.25in,clip,keepaspectratio]{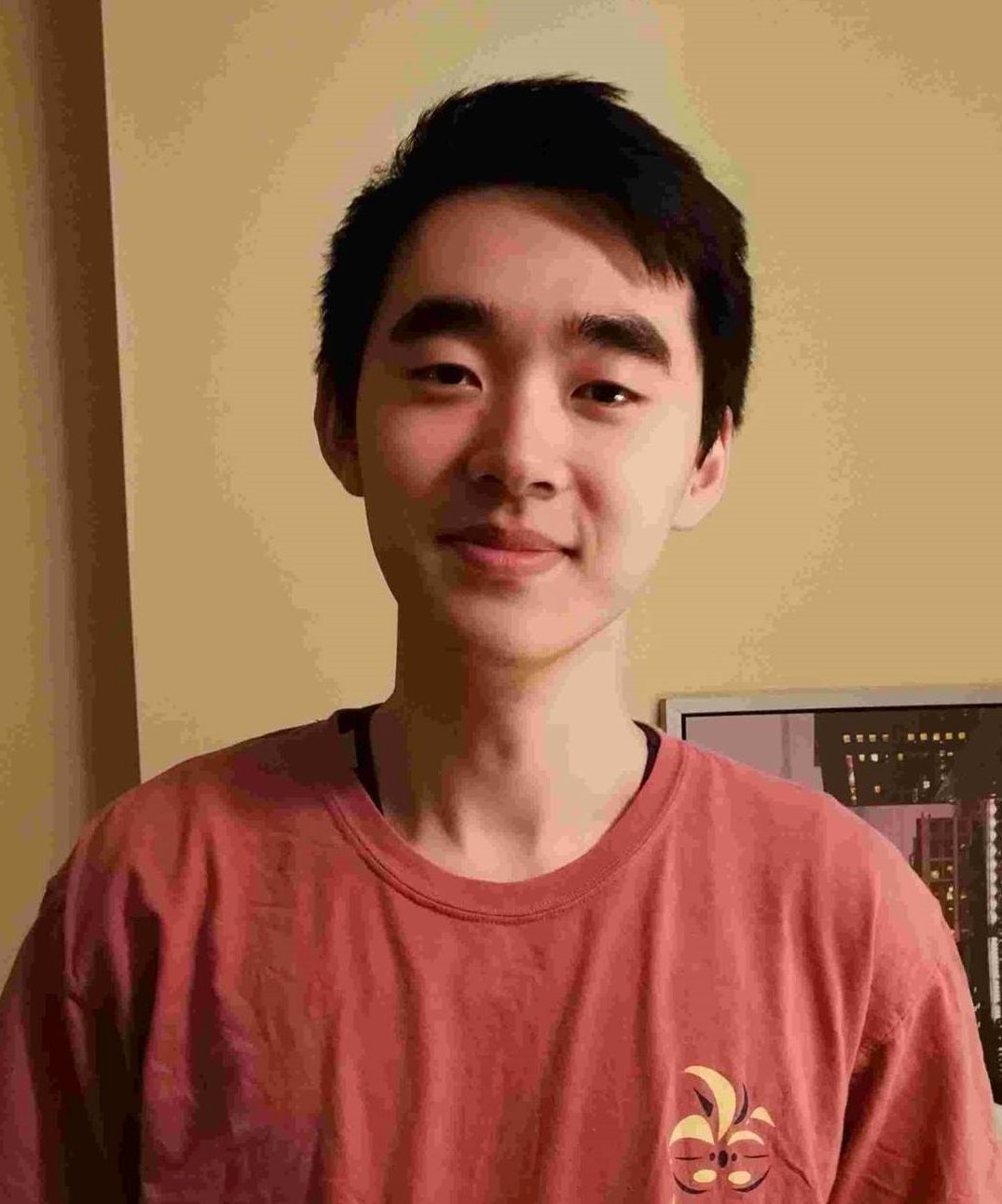}}]{Fanjun Bu} is a Ph.D. candidate in the Department of Computer Science at the Cornell University. He received his B.S. degree in Applied Math and Computer Science from the Johns Hopkins University. His research interest includes human-robot interaction, human-computer interaction, and cognitive robotics.

\end{IEEEbiography}

\begin{IEEEbiography}[{\includegraphics[width=1in,height=1.25in,clip,keepaspectratio]{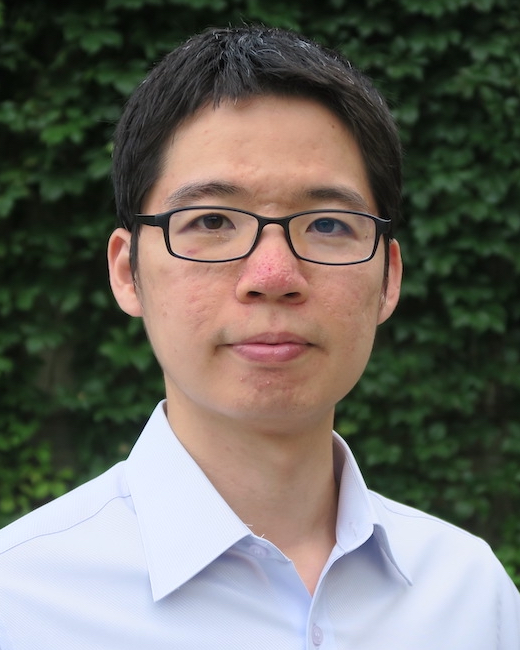}}]{Chien-Ming Huang} is a John C. Malone Assistant Professor in the Department of Computer Science and the Malone Center for Engineering in Healthcare at the Johns Hopkins University. His research focuses on designing and developing interactive systems that provide physical, social, and cognitive support to people with diverse needs. Dr. Huang completed his postdoctoral training at Yale University, his PhD in Computer Science at the University of Wisconsin--Madison, and his MS in Computer Science at the Georgia Institute of Technology. His research has received media coverage from MIT Technology Review, Tech Insider, and Science Nation.
\end{IEEEbiography}

\EOD

\end{document}